# N-dimensional nonlinear prediction with MLP

*Marcos Faúndez-Zanuy*

Escola Universitària Politècnica de Mataró
Universitat Politècnica de Catalunya (UPC)
Avda. Puig i Cadafalch 101-111, E-08303 Mataró (BARCELONA) SPAIN
`faundez@eupmt.es`

**Abstract**

In this paper we propose a Non-Linear Predictive Vector quantizer (PVQ) for speech coding, based on Multi-Layer Perceptrons. With this scheme we have improved the results of our previous ADPCM coder with nonlinear prediction, and we have reduced the bit rate up to 1 bit per sample.

## 1. Introduction

In [1] we proposed a scheme for nonlinear vectorial predictor based on neural nets. In [2] we applied the predictor for speech coding. This scheme is known as Non-Linear Predictive Vector Quantization [3, chap. 13] NL-PVQ. This system is similar to an ADPCM speech coder, where the NL predictor replaces the LPC predictor in order to obtain an ADPCM scheme with non-linear vectorial prediction. In addition, the scalar quantizer is replaced by a vectorial quantizer.

We have checked that increasing the dimension of the predicted vector it is possible to increase the SEGSNR and to extend the operating range to lower bit rates.

## 2. Vectorial nonlinear prediction

Our nonlinear predictor consists on a Multi Layer Perceptron (MLP) with 10 inputs, 2 neurons in the hidden layer, and $N$ outputs, where $N$ is the dimension of the vectors (see figure 1). In this paper we use $N=1,2..6$. The selected training algorithm is the Levenberg-Marquardt, that computes the approximate Hessian matrix, because it is faster and achieves better results than the classical back-propagation algorithm. We also apply a multi-start algorithm with five random initializations for each neural net. In [4] we studied several training schemes, and we concluded that the most suitable is the combination between Bayesian regularization and a committee of neural nets (each neural net is the result of training one random initialization).

We have checked the vectorial prediction in several scenarios:
1. Scalar prediction and scalar quantization: this scheme is equivalent to a neural net trained with hints ($N$ outputs are used during training phase, but only the first output is used for prediction). Thus, we train a vectorial predictor but we use it as a scalar predictor.
2. Vectorial prediction and scalar quantization: all the neural net outputs are used for prediction, but the adaptive scalar quantizer based on multipliers [6] is used consecutively in order to quantize the $N$ output prediction errors. Although this quantizer has been tuned up for linear predictors, we have found in our previous work that it is also suitable for nonlinear prediction and it is able to remove the first order dependencies between consecutive samples [3].
3. Vectorial prediction and vector quantization: same situation than the previous scenario, but the scalar quantizer is replaced by a VQ. In [3] we obtain that this scheme with $N=2$ was unable to outperform the scalar quantizer, and that first order dependencies exist. In that paper we conclude that the system should be improved with VQ memory quantizer or increasing the vector dimension. In this paper, we have studied the results for higher vector dimensions ($N>2$) and we have improved the SEGSNR with smaller bit rates.

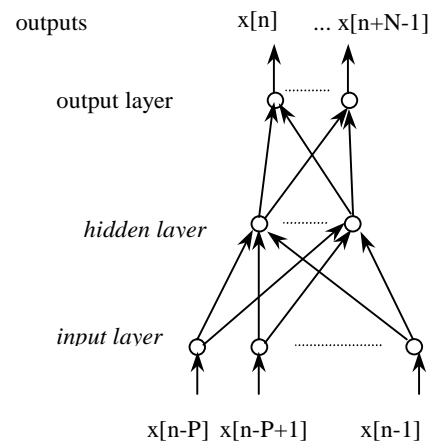

Figure 1. Vectorial predictor based on a MLP

### 2.1. Experiment conditions

We have used the same database than in our previous papers [1],[2],[4], which consists on 8 speakers (4 male and 4 female). The number of inputs is $P=10$, and the number of outputs is variable in the range $1 \leq N \leq 6$. The frame length is 200 samples. In order to obtain always the same number of training patterns, the following input/output patterns have been used:

For i=0:frame_length−1,
$$input(i) = \{x[n+i-P], \cdots, x[n+i-1],\}$$
$$output(i) = \{x[n+i], \cdots x[n+i+N-1]\}$$
end

Thus, there is a shift of one sample between consecutive input patterns during the training of the neural net (if the shift would be *N* samples, the number of training patters would decrease and couldn't be enough for high values of *N*.

Obviously we have slightly modified the frame length for *N*=3 (201 samples) and *N*=6 (204 samples), in order to achieve an exact division of frame_length by *N*.

On the other hand, the shift between consecutive input patterns when the neural net is acting as a predictor is equal to *N*.

This is a backward-adaptive ADPCM scheme. Thus, the coefficients are computed over the previous frame, and it is not needed to transmit the coefficients of the predictor, because the receiver has already decoded the previous frame and can obtain the same set of coefficients.

In [1] we showed that the computation of a vectorial predictor based on a MLP is not critical. This has been checked using the scenario number 1 of section 2. In this situation the vectorial prediction training procedure can be interpreted as a particular case of neural net training with output hints. We obtained similar performance than a neural net with only one output neuron and same number of neurons on the input and hidden layers. Of course, consecutive samples are highly correlated, so really the neural net is not bounded to learn a significative amount of "new information". Thus, the generalization of the scalar NL predictor to a NL vectorial prediction does not imply a great difference with respect to the scalar predictor.

## 3. Vectorial quantizer

In order to design a vectorial quantizer (VQ) it is need a training sequence. The optimal design procedure must be iterative [4], because in a PVQ scheme the VQ is inside the loop of the ADPCM scheme. In order to achieve a "universal VQ", it should be obtained with as many speakers and sentences as possible and evaluated with a different database. We have used only one speaker for VQ generation and 7 different speakers for PVQ system evaluation. We have used two different methods for codebook generation given a training sequence: random initialization plus the generalized Lloyd iteration, and the LBG algorithm [2].

We have used the following procedure:

1. A speech database is PVQ coded with a vectorial predictor and an adaptive scalar quantizer based on multipliers [6]. Although the prediction algorithm is vectorial, the residual error is scalar quantized, applying the scalar quantizer consecutively to each component of the residual vector.
2. We have used the residual signal of one sentence uttered by a female speaker (approximately 20000/*N* vectors) and 3 quantization bits (Nq=3) as a training sequence.
3. A codebook is designed for VQ sizes in the range *Nq*=[5, 9]. Thus, the bit rate is *Nq*/*N* bits/sample.

It is interesting to observe that although the same sentence would be used two times the prediction error would be different, due to the random initialization of the neural net weights. This does not happen with the linear prediction coefficients, because they are obtained with a deterministic procedure.

## 4. Results

This section summarizes the results for the different scenarios proposed in section 2.

### 4.1. Scalar prediction and scalar quantization

This situation is equivalent to a neural net trained with hints on the output. This experiment is interesting in order to evaluate the ability of the neural net to work fine when the number of output neurons is increased.

Table one shows the results for several combinations of *N* and *Nq*, where *N* is the dimension of the output predicted vectors, and *Nq* is the number of quantization bits of the scalar quantizer.

Table 1: SEGSNR with scalar prediction & quantizer.

| Nq | N=1 | | N=2 | | N=3 | | N=4 | | N=5 | | N=6 | |
|---|---|---|---|---|---|---|---|---|---|---|---|---|
| | m | σ | m | σ | m | σ | m | σ | m | σ | m | σ |
| 2 | 14.53 | 4.9 | 14.20 | 5.1 | 13.96 | 5.0 | 13.76 | 5 | 13.51 | 5 | 13.34 | 5.1 |
| 3 | 20.55 | 5.9 | 20.35 | 5.8 | 20.22 | 5.8 | 19.97 | 5.7 | 19.77 | 5.6 | 19.46 | 5.5 |
| 4 | 25.78 | 6.5 | 25.55 | 6.3 | 25.13 | 6.3 | 24.98 | 6.3 | 24.81 | 6.1 | 24.55 | 6.1 |
| 5 | 30.45 | 6.8 | 30.35 | 6.7 | 30.08 | 6.5 | 29.67 | 6.0 | 29.58 | 6.3 | 29.41 | 6.3 |

We observe that there is a slight degradation on the SEGSNR when N is increased. We will check in the next sections that this effect can be compensated by the improvement that introduces the vectorial quantizer.

### 4.2. Vectorial prediction and scalar quantization

We apply the adaptive scalar quantizer based on multipliers successively to the different vectorial predictor outputs. Table 2 summarizes the results for several combinations of *N* and *Nq*.

We can observe that the reduction on SEGSNR is greater than in the previous scenario. Thus, the scalar quantizer can not to take advantage of the vectorial prediction, and a different quantization scheme must be evaluated.

Table 2: SEGSNR with vectorial Prediction and scalar quantization.

| Nq | N=2 | | N=3 | | N=4 | | N=5 | | N=6 | |
|---|---|---|---|---|---|---|---|---|---|---|
| | m | σ | M | σ | m | σ | m | σ | m | σ |
| 2 | 12.86 | 4.5 | 11.65 | 4.10 | 11.21 | 3.93 | 10.68 | 3.87 | 10.38 | 3.67 |
| 3 | 18.61 | 5.17 | 17.43 | 4.73 | 16.95 | 4.34 | 16.57 | 4.25 | 16.13 | 4.24 |
| 4 | 23.25 | 5.33 | 22.3 | 4.97 | 21.98 | 4.55 | 21.74 | 4.35 | 21.50 | 4.47 |
| 5 | 27.7 | 5.52 | 26.96 | 5.13 | 26.81 | 4.82 | 26.58 | 4.62 | 26.22 | 4.71 |

### 4.3. Vectorial prediction and vectorial quantization

This situation corresponds to the Non-linear vectorial predictor with a vectorial quantizer, where *N* is the dimension of the output predicted vectors and *Nq* is the number of bits of the codebook. Thus, the bit rate is *Nq/N*.

Table 3 summarizes the results for several values of *Nq* and *N*. In order to compare the SEGSNR at the different bit rates, we have ploted all the results in a same figure (see fig. 2).

Table 3: SEGSNR with vectorial prediction & quantizer.

| N | Nq=5 | | Nq=6 | | Nq=7 | | Nq=8 | | Nq=9 | |
|---|---|---|---|---|---|---|---|---|---|---|
| | m | σ | m | σ | m | σ | m | σ | m | σ |
| 2 | 15.8 | 6.9 | 19 | 5.9 | 21.4 | 6.1 | 25 | 5.7 | | |
| 3 | 9.21 | 6.61 | 12.14 | 5.72 | 14.63 | 5.05 | 16.77 | 4.88 | 18.34 | 4.87 |
| 4 | 7.81 | 6.02 | 10.16 | 5.91 | 12.15 | 4.89 | 13.50 | 4.53 | 14.99 | 4.68 |
| 5 | 8.52 | 6.17 | 10.70 | 5.51 | 12.37 | 5.02 | 13.83 | 5.08 | 14.38 | 5.32 |
| 6 | 6.96 | 4.79 | 8.14 | 4.49 | 9.63 | 4.24 | 10.46 | 4.29 | 11.42 | 4.58 |

It is important to take into account that although it seems that for a given *Nq* the SEGSNR is reduced when *N* is increased, the bit rate is also reduced, because the number of bits of each codeword must be split by the vector dimension *N*. Thus, for a given bit rate, the SEGSNR is higher if *N* is increased (see figure 2).

## 5. Study of the quantizer

In order to study the quantizer, we propose to evaluate the zero order entropy $H_0(X)$ and the first order entropy $H_1(X)$ of the codewords, where:

- $H_0(x) = \sum_{i=1}^{M} P_i \log_2 \frac{1}{P_i}$

- $H_1(X) = \sum_{j=1}^{M} \sum_{i=1}^{M} P(ij) \log_2 \frac{1}{P(i|j)}$

- $P_i$ is the probability of the codeword *i*.
- $P(i|j)$ is the probability of the codeword *i* knowing that the previous codeword has been the codeword *j*.

It is important to take into account that this formulation is valid for scalar and vectorial quantization. The unique difference is that in the former case each codeword is equivalent to one sample, while in the latest one each codeword is equivalent to a vector (group of samples).

It would be interesting to study higher order entropies, but the amount of required data and the computational burden makes this evaluation unpractical.

The better designed the quantizer, the higher the entropy, because all the codewords have the same probability of being chosen. In this case, $H_0(X) \cong N_q$.

Otherwise, the outputs of the quantizer (codewords) can be encoded with a lossless method (for example Huffmann) in order to reduce the data rate.

On the other hand, if $H_1(X) << H_0(X)$ means that there is a strong correlation between consecutive quantizer outputs, and two observations can be made:

1. The outputs of the quantizer (codewords) can be encoded with a lossless method (for example Huffmann) in order to reduce the data rate.
2. The quantizer can be improved taking into account the previous sample (using a memory quantizer). The goal is to obtain $H_1(X) \cong H_0(X) \cong N_q$ (remember that $H_1(X) \leq H_0(X) \leq N_q$ by definition). In this case, all the codewords are equal probability used, and $P(x[n]|x[n-1]) \cong P(x[n])$, so the quantizer has removed the first order dependencies, and no improvement is achieved by means of a Huffmann code.

Our goal is the latest observation, rather than the former one, because the better the quantizer, the better the prediction (both systems are in a closed loop). If the entropy is smaller than *Nq* means that some codewords are not used, so the useful number of quantization bits is smaller than *Nq*.

Table 4 shows the results using one sentence of the database. Special care must be taken in order to obtain a good estimation of the probabilities and conditional probabilities, because the number of different codewords is $2^{N_q}$, and the higher the number of different possible codewords, the higher the number of codewords we need to obtain a good estimation of the probabilities of these codewords. Specially for the first order entropy, because the number of different combinations is $2^{N_q} \times 2^{N_q} = 2^{2N_q}$. In [2] we experimentally shown than it is easy to compute the zero and first order entropies up to *Nq*=5 (with a significative increase on the number of codewords used to compute the statistics did not imply a modification of the results). Thus, it is important to take into account that $H_1$ values are underestimated for *Nq*>5 due to the limited amount of samples used to compute $P(i|j)$.

Table 4. $H_0$ and $H_1$ of the codewords

| N | Nq=5 | | Nq=6 | | Nq=7 | | Nq=8 | | Nq=9 | |
|---|---|---|---|---|---|---|---|---|---|---|
| | Ho | $H_1$ | Ho | $H_1$ | Ho | $H_1$ | Ho | $H_1$ | Ho | $H_1$ |
| 3 | 3.98 | 3.23 | 5.13 | 3.81 | 6.29 | 3.91 | 7.26 | 3.6 | 8.02 | 3.06 |
| 4 | 3.87 | 3.04 | 5.01 | 3.66 | 6.43 | 3.96 | 7.37 | 3.55 | 8.09 | 3.04 |
| 5 | 4.25 | 3.32 | 5.49 | 3.95 | 6.73 | 3.84 | 7.73 | 3.35 | 8.48 | 2.67 |
| 6 | 4.40 | 3.48 | 5.54 | 3.91 | 6.57 | 3.89 | 7.34 | 3.50 | 8.08 | 2.96 |

## 6. Conclusion

In this paper we have proposed a Non-linear Predictive Vector Quantization speech encoder based on a multilayer perceptron. We think that this scheme could be a preliminar step towards a more sophisticated coder like the CELP scheme.

We have also proposed a methodology for the analysis of a quantizer, based on two propositions:
1. If the quantizer is well designed, the zero order entropy will be approximately equal to the number of quantization bits (that is, all the quantization outputs have the same probability).
2. If the quantizer exploits the correlation between consecutive outputs, the first order entropy will be approximately the same than the zero order entropy. Otherwise, it will be smaller.

In our previous paper [4] we had shown that an ADPCM scheme with scalar nonlinear predictor outperforms the same scheme with scalar linear predictor. In this paper we have shown that a vectorial nonlinear predictor outperforms a scalar nonlinear predictor for $N$=5. This result is analogous to the reported in [5].

On the other hand, we have reduced the bitrate up to 1 bit per sample, while the classical scalar ADPCM scheme is over 2 bits per sample.

### Acknowledgement

This work has been supported by the Spanish grant CICYT TIC2000-1669-C04-02

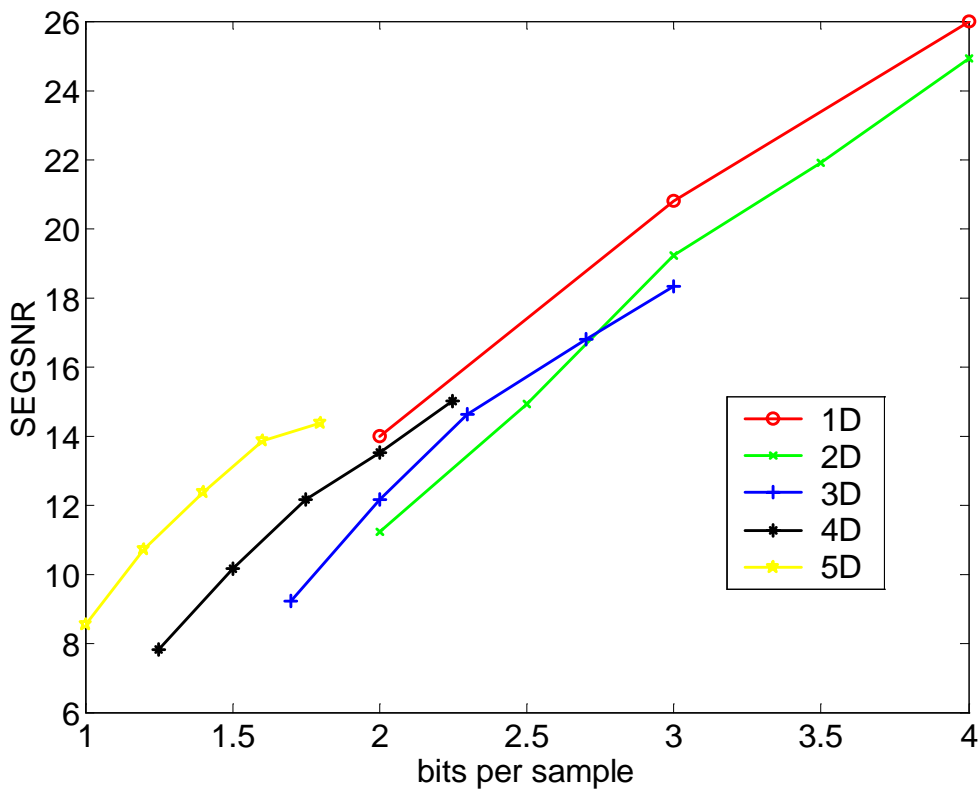

Figure 2: SEGSNR as function of the bitrate.